\documentclass[11pt,a4paper]{article}
\usepackage[hyperref]{acl2019}
\usepackage{times}
\usepackage{latexsym}
\usepackage{url}

\usepackage{amsmath}
\usepackage[pdftex]{graphicx}

\usepackage{tabularx}
\usepackage{bm}
\usepackage{color}
\usepackage{subfig}

\aclfinalcopy 
\def\aclpaperid{478} 

\title{Unsupervised Neural Single-Document Summarization of Reviews \\ via Learning Latent Discourse Structure and its Ranking}

\author{Masaru Isonuma${}^{1}$ \qquad \textbf{Junichiro Mori}${}^{1,2}$ \qquad \textbf{Ichiro Sakata}${}^{1}$ \\
		${}^{1}$ \textbf{The University of Tokyo} \qquad ${}^{2}$ \textbf{RIKEN}\\
  		\texttt{\{isonuma, isakata\}@ipr-ctr.t.u-tokyo.ac.jp} \\ 
  		\texttt{mori@mi.u-tokyo.ac.jp} \\
  		}
  	
\date{}

\begin{document}
\maketitle
\begin{abstract}
This paper focuses on the end-to-end abstractive summarization of a single product review {\em without supervision}. We assume that a review can be described as a {\em discourse tree}, in which the summary is the root, and the child sentences explain their parent in detail. By recursively estimating a parent from its children, our model learns the latent discourse tree without an external parser and generates a concise summary. We also introduce an architecture that {\em ranks} the importance of each sentence on the tree to support summary generation focusing on the main review point. The experimental results demonstrate that our model is competitive with or outperforms other unsupervised approaches. In particular, for relatively long reviews, it achieves a competitive or better performance than supervised models. The induced tree shows that the child sentences provide additional information about their parent, and the generated summary abstracts the entire review.

\end{abstract}

\begin{figure*}[t!]
\begin{center}
\includegraphics[width=15cm]{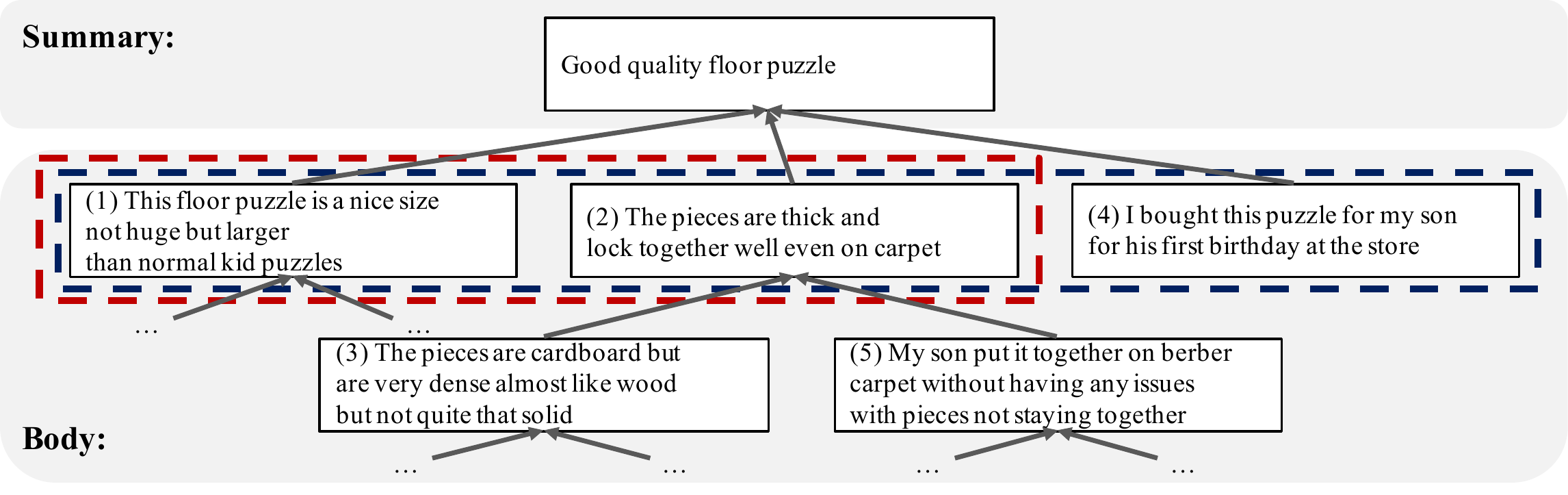}
\end{center}
\caption{Example of the discourse tree of a jigsaw puzzle review. \textcolor{blue}{StrSum} induces the latent tree and generates the summary from the children of a root, while \textcolor{red}{DiscourseRank} supports it to focus on the main review point.}
\label{fig:introduction}
\end{figure*}

\section{Introduction}
  
The need for automatic document summarization is widely increasing because of the vast amounts of online textual data that continue to grow. As for product reviews on E-commerce websites, succinct summaries allow both customers and manufacturers to obtain large numbers of opinions \cite{liu2012survey}. Under these circumstances, supervised neural network models have achieved wide success, using a large number of reference summaries \cite{wang2016neural, ma2018hierarchical}. However, a model trained on these summaries cannot be adopted in other domains, as salient phrases are not common across domains. It requires a significant cost to prepare large volumes of references for each domain \cite{isonuma2017extractive}.

An unsupervised approach is a possible solution to such a problem. Previously, unsupervised learning has been widely applied to extractive approaches \cite{radev2004centroid, mihalcea2004textrank}. As mentioned in \cite{carenini2013multi, gerani2014abstractive}, extractive approaches often fail to provide an overview of the reviews, while abstractive ones successfully condense an entire review via paraphrasing and generalization. Our work focuses on the one-sentence abstractive summarization of a single-review without supervision.

The difficulties of unsupervised abstractive summarization are two-fold: obtaining the representation of the summaries, and learning a language model to decode them. As an unsupervised approach for multiple reviews, \citet{chu2018unsupervised} regarded the mean of the document embeddings as the summary, while learning a language model via the reconstruction of each review. By contrast, such an approach cannot be extended to a single-review directly, because it also condenses including trivial or redundant sentences (its performance is demonstrated in Section \ref{sec:result}). 

To overcome these problems, we apply the {\em discourse tree} framework. Extractive summarization and document classification techniques sometimes use a discourse parser to gain a concise representation of documents \cite{hirao2013single, bhatia2015better, ji2017neural}; however, \citet{ji2017neural} pointed out the limitations of using external discourse parsers. In this context, \citet{liu2018learning} proposed a framework to induce a latent discourse tree without a parser. While their model constructed the tree via a supervised document classification task, our model induces it by identifying and reconstructing a parent sentence from its children. Consequently, we gain the representation of a summary as  the root of the induced latent discourse tree, while learning a language model through reconstruction.

Figure \ref{fig:introduction} shows an example of a jigsaw puzzle review and its dependency-based discourse tree. The summary describes its quality. The child sentences provide an explanation in terms of the size ($1$st) and thickness ($2$nd), or provide the background ($4$th). Thus, we assume that reviews can generally be described as a multi-root non-projective discourse tree, in which the summary is the root, and the sentences construct each node. The child sentences present additional information about the parent sentence.

To construct the tree and generate the summary, we propose a novel architecture; {\it StrSum}. It reconstructs a parent from its children recursively and induces a latent discourse tree without a parser. As a result, our model generates a summary from the {\em surrounding sentences of the root} while learning a language model through reconstruction in an end-to-end manner. We also introduce {\it DiscourseRank}, which ranks the importance of each sentence in terms of the number of {\em descendants}. It supports StrSum to generate a summary that focuses on the main review point. 

The contributions of this work are three-fold:
\vspace{-0.35\baselineskip} 
\begin{itemize}
\setlength{\itemsep}{0.05pt}
  \item We propose a novel unsupervised end-to-end model to generate an abstractive summary of a single product review while inducing a latent discourse tree
  \item The experimental results demonstrate that our model is competitive with or outperforms other unsupervised models. In particular, for long reviews, it achieves a competitive or better performance than the supervised models.
  \item The induced tree shows that the child sentences present additional information about their parent, and the generated summary abstracts for the entire review. 
\end{itemize}
\vspace{-0.5\baselineskip}

\begin{figure*}[htb]
\begin{center}
\includegraphics[width=15.5cm]{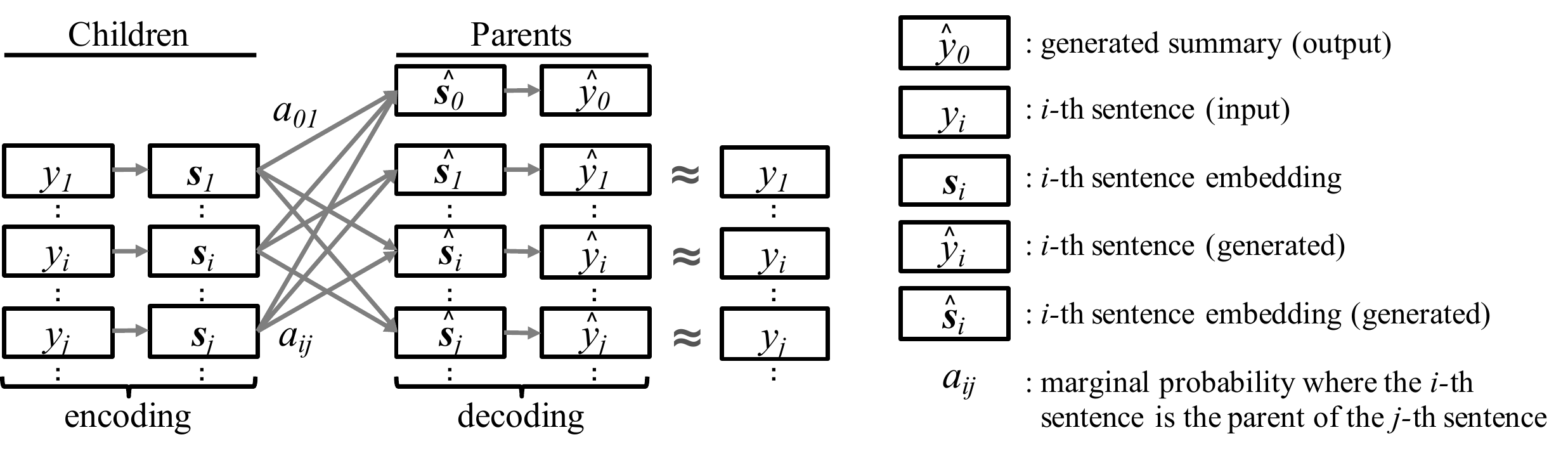}
\end{center}
\caption{Outline of StrSum.}
\label{fig:proposedmodel}
\end{figure*}

\section{Proposed Model}

In this section, we present our unsupervised end-to-end summarization model with descriptions of StrSum and DiscourseRank.

\subsection{StrSum: Structured Summarization}

\textbf{Model Training:} The outline of StrSum is presented in Figure \ref{fig:proposedmodel}. $y_i$ and ${\bm s}_i \in \mathcal{R}^{d}$ indicate the $i$-th sentence and its embedding in a document $D = \{y_1, \ldots, y_n\}$, respectively. $w^t_i$ is the $t$-th word in a sentence $y_i = \{w^1_i, \ldots, w^l_i\}$. ${\bm s}_i$ is computed via a max-pooling operation across hidden states ${\bm h}^t_i \in \mathcal{R}^{d}$ of the Bi-directional Gated Recurrent Units (Bi-GRU):
\begin{align}
\overrightarrow{\bm h}^t_i &= \overrightarrow{\rm GRU}(\overrightarrow{\bm h}^{t-1}_i, w^t_i) \\
\overleftarrow{\bm h}^t_i &= \overleftarrow{\rm GRU}(\overleftarrow{\bm h}^{t+1}_i, w^t_i) \\
{\bm h}^t_i &= [\overrightarrow{\bm h}^t_i, \overleftarrow{\bm h}^t_i] \\
\forall m \in \{1, \ldots, d\}, \ {\bm s}_{i, m} &= {\displaystyle \max_{t}} \ {\bm h}^t_{i, m}
\end{align}

Here, we assume that a document $D$ and its summary compose a {\em discourse tree}, in which the root is the summary, and all sentences are the nodes. We denote $a_{ij}$ as the {\em marginal probability of dependency} where the $i$-th sentence is the parent node of the $j$-th sentence. In particular, $a_{0j}$ denotes the probability that a root node is the parent (see Figure \ref{fig:proposedmodel}). We define the probability distribution $a_{ij} \ (i \in \{0, \ldots, n\}, j \in \{1, \ldots, n\})$ as the posterior marginal distributions of a non-projective dependency tree. The calculation of the marginal probability is explained later. 

Similar to \cite{liu2018learning}, to prevent overload of the sentence embeddings, we decompose them into two parts: 
\begin{align}
[{\bm s}_{i}^e, {\bm s}_{i}^f] = {\bm s}_{i}
\end{align}
where the semantic vector ${\bm s}_{i}^e \in \mathcal{R}^{d_e}$ encodes the semantic information, and the structure vector ${\bm s}_{i}^f \in \mathcal{R}^{d_f}$ is used to calculate the marginal probability of dependencies.

The embedding of the parent sentence $\hat{\bm s}_i$ and that of the {\em summary} $\hat{\bm s}_{0}$ are defined with parameters ${\bm W}_s \in \mathcal{R}^{d_e*d_e}$ and ${\bm b}_s \in \mathcal{R}^{d_e}$ as:
\begin{align}
\hat{\bm s}_{i} = \tanh \Bigl\{ {\bm W}_s ({\displaystyle \sum_{j=1}^{n} \ a_{ij}}  {\bm s}_{j}^e) + {\bm b}_s \Bigl\}
\end{align}

Using $\hat{\bm s}_i$, the GRU-decoder learns to reconstruct the $i$-th sentence, \emph{i.e.}, to obtain the parameters ${\bm \theta}$ that maximize the following log likelihood:
\begin{eqnarray}
\label{eq:likelihood}
\sum_{i=1}^{n} \sum_{t=1}^{l} \log{P(w_i^t| w_i^{<t}, \hat{\bm s}_i, \bm \theta)}
\end{eqnarray}

\textbf{Summary Generation:} An explanation of how the training contributes to the learning of a language model and the gaining of the summary embedding is provided here. As for the former, the decoder learns a language model to generate grammatical sentences by reconstructing the document sentences. Therefore, the model can appropriately decode the summary embedding to $\hat{y}_0$. 

As for the latter, if the $j$-th sentence contributes to generating the $i$-th one, $a_{ij}$ get to be higher. This mechanism models our assumption that {\em child sentences can generate their parent sentence, but not vice versa}, because the children present additional information about their parent. Hence, the most concise $k$-th sentences (\emph{e.g.}, the $1$st, $2$nd, and $4$th in Figure \ref{fig:introduction}), provide less of a contribution to the reconstruction of any other sentences. Thus, $a_{ik}$ get to be lower for $\forall i: i\neq0$. Because $a_{ik}$ satisfies the constraint $\sum_{i=0}^{n} a_{ik} \! = \! 1$, $a_{0k}$ is expected to be larger, and thus the $k$-th sentence contributes to the construction of the summary embedding $\hat{\bm s}_0$.

\textbf{Marginal Probability of Dependency:} The calculation of the marginal probability of dependency, $a_{ij}$, is explained here. We first define the weighted adjacency matrix ${\bm F} = (f_{ij}) \in \mathcal{R}^{(n+1)*(n+1)}$, where the indices of the first column and row are $0$, denoting the root node. $f_{ij}$ denotes the un-normalized weight of an edge between a parent sentence $i$ and its child $j$. We define it as a pair-wise attention score following \cite{liu2018learning}. By assuming a {\em multi\--root} discourse tree, $f_{ij}$ is defined as:
\begin{align}
f_{ij} &=
\begin{cases}
\exp({\bm w}_r^\top {\bm s}_j^f) & \! (i = 0 \land j \geq 1) \\
\exp({\bm p}_i^\top {\bm W}_f {\bm c}_j) &\! (i \geq 1 \land j \geq 1 \land i \neq j) \\
0 &\! (j = 0 \lor i = j) \\
\end{cases} \\
{\bm p}_i &= \tanh({{\bm W}_p {\bm s}_i^f} + {\bm b}_p) \\
{\bm c}_j &= \tanh({{\bm W}_c {\bm s}_j^f} + {\bm b}_c)
\end{align}
where ${\bm W}_f \in \mathcal{R}^{d_f*d_f}$ and ${\bm w}_r \in \mathcal{R}^{d_f}$ are parameters for the transformation. ${\bm W}_p \in \mathcal{R}^{d_f*d_f}$ and ${\bm b}_p \in \mathcal{R}^{d_f}$ are the weight and bias respectively, for constructing the representation of the parent nodes. ${\bm W}_c \in \mathcal{R}^{d_f*d_f}$ and ${\bm b}_c \in \mathcal{R}^{d_f}$ correspond to those of the child nodes.

We normalize $f_{ij}$ into $a_{ij}$ based on \cite{koo2007structured}. $a_{ij}$ corresponds to the proportion of the total weight of the spanning trees containing an edge $(i,j)$:
\begin{align}
a_{ij}(\bm F) &= \frac{\sum_{t \in T:(i,j) \in t} {v(t| {\bm F})}}{\sum_{t \in T} {v(t| {\bm F})}} \\
&= \frac{\partial \log Z({\bm F})}{\partial f_{ij}} \label{eq:probability} \\
v(t|{\bm F}) &= \prod_{(i, j) \in t} f_{ij} \label{eq:treescore} \\
Z({\bm F}) &= \sum_{t \in T} {v(t| {\bm F})}
\end{align}
where $T$ denotes the set of all spanning trees in a document $D$. $v(t|{\bm F})$ is the weight of a tree $t \in T$, and $Z({\bm F})$ denotes the sum of the weights of all trees in $T$. From the Matrix-Tree Theorem \cite{tutte1984graph}, $Z({\bm F})$ can be rephrased as:
\begin{eqnarray}
Z({\bm F}) = |L_{0}(\bm F)|
\end{eqnarray}
where $L({\bm F}) \in \mathcal{R}^{(n+1)*(n+1)}$ and $L_{0}(\bm F) \in \mathcal{R}^{n*n}$ are the Laplacian matrix of ${\bm F}$ and its principal submatrix formed by deleting row $0$ and column $0$, respectively. By solving Eq. \ref{eq:probability}, $a_{ij}$ is given by:
\begin{align}
\label{eq:marginalprobability}
a_{0j} &= f_{0j} \left[L_{0}^{-1}(\bm F)\right]_{jj} \\ 
a_{ij} &= f_{ij} \left[L_{0}^{-1}(\bm F)\right]_{jj} - f_{ij} \left[L_{0}^{-1}(\bm F)\right]_{ji}
\end{align}
 
\subsection{DiscourseRank}

StrSum generates the summary under the large influence of the child sentences of the root. Therefore, sentences that are not related to the rating (\emph{e.g.}, the $4$th in Figure \ref{fig:introduction}) also affect the summary and can be considered noise. Here, we assume that meaningful sentences (\emph{e.g.}, the $1$st and $2$nd in Figure \ref{fig:introduction}) typically have more {\em descendants}, because many sentences provide the explanation of them. Hence, we introduce the {\it DiscourseRank} to rank the importance of the sentences in terms of the number of descendants. Inspired by PageRank \cite{page1999pagerank}, the DiscourseRank of the root and $n$ sentences at the $t$-th iteration ${\bm r^t} = \left[r_0, \ldots, r_n \right] \in \mathcal{R}^{(n+1)}$ is defined as:
\begin{align}
{\bm r}^{t+1} &= \lambda {\bm \hat{A}}{\bm r}^{t} + (1-\lambda){\bm v} \label{eq:discourserank}\\
\hat{a}_{ij} &=
\begin{cases}
0 & (i = 0 \land j = 0)\\
\frac{1}{n} & (i \geq 1 \land j = 0) \\
a_{ij} & (j \geq 1) \\
\end{cases}
\end{align}
\noindent
where $\hat{\bm A} = (\hat{a}_{ij}) \in \mathcal{R}^{(n+1)*(n+1)}$ denotes the stochastic matrix for each dependency, $\lambda$ is a damping factor, and ${\bm v} \in \mathcal{R}^{(n+1)}$ is a vector with all elements equal to $1/(n+1)$. Eq.\ref{eq:discourserank} implies that $r_i$ reflects $r_j$ more if the $i$-th sentence is more likely to be the parent of the $j$-th sentence. The $\bm r$ solution and updated score of the edge $(0, j)$ $\bar{a}_{0j} \ (j \in \{1, \ldots, n\})$ are calculated by:
\begin{align}
{\bm r} &= (1-\lambda)(I - \lambda{\bm \hat{A}})^{-1}{\bm v}\\
\bar{a}_{0j} &= a_{0j} r_j
\end{align}
The updated score $\bar{a}_{0j}$ is used to calculate the summary embedding $\hat{s}_0$ instead of Eq.\ref{eq:marginalprobability}. As a result, the generated summary reflects the sentences with a higher marginal probability of dependency on the root, while focusing on the main review point.

\section{Related work}
\subsection{Supervised Review Summary Generation}

Several previous studies have addressed abstractive summarization for product reviews \cite{carenini2013multi, di2014hybrid, bing2015abstractive, yu2016product}; however, their output summaries are not guaranteed to be grammatical \cite{wang2016neural}. Neural sequence-to-sequence models have improved the quality of abstractive summarization. Beginning with the adaptation to sentence summarization \cite{rush2015neural, chopra2016abstractive}, several studies have tackled the generation of an abstractive summary of news articles \cite{nallapati2016abstractive, see2017get, tan2017abstractive, paulus2018deep}. With regard to product reviews, the neural sequence-to-sequence based model \cite{wang2016neural} and joint learning with sentiment classification \cite{ma2018hierarchical, wang2018self} have improved the performance of one-sentence summarization. Our work is also based on the neural sequence-to-sequence model, while introducing the new concept of generating the summary by recursively reconstructing a parent sentence from its children.

\subsection{Unsupervised Summary Generation}

Although supervised abstractive summarization has been successfully improved, unsupervised techniques have still not similarly matured. \citet{ganesan2010opinosis} proposed {\it Opinosis}, a graph-based method for generating review summaries. Their method is word-extractive, rather than abstractive, because the generated summary only contains words that appear in the source document. With the recently increasing number of neural summarization models, \citet{miao2016language} applied a variational auto-encoder for semi-supervised sentence compression. \citet{chu2018unsupervised} proposed {\it MeanSum}, an unsupervised neural multi-document summarization model for reviews. However, their model is not aimed at generating a summary from a single document and could not directly be extended. Although several previous studies \cite{fang2016proposition, Dohare2018UnsupervisedSA} have used external parsers for unsupervised abstractive summarization, our work, to the best of our knowledge, proposes the first unsupervised abstractive summarization method for a single product review that does not require an external parser. 

\subsection{Discourse Parsing and its Applications}

Discourse parsing has been extensively researched and used for various applications. \citet{hirao2013single, kikuchi2014single, yoshida2014dependency} transformed a rhetorical structure theory-based discourse tree (RST-DT; \citealp{mann1988rhetorical}) into a dependency-based discourse tree and regarded the root and the surrounding elementary discourse units as a summary. \citet{gerani2014abstractive} constructed a discourse tree and ranked the aspects of reviews for summarization. \citet{bhatia2015better,ji2017neural} also constructed a dependency-based discourse tree for document classification. \citet{ji2017neural} pointed out the limitations of using external parsers, demonstrating that the performance depends on the amount of the RST-DT and the domain of the documents. 

Against such a background, \citet{liu2018learning} proposed a model that induces a latent discourse tree without an external corpus. Inspired by structure bias \cite{cheng2016neural, kim2017structured}, they introduced {\it Structured Attention}, which normalizes attention scores as the posterior marginal probabilities of a non-projective discourse tree. The probability distribution of Structured Attention implicitly represents a discourse tree, in which the child sentences present additional information about their parent. We extend it to the unsupervised summarization, \emph{i.e.}, obtaining a summary as the root sentence of a latent discourse tree. While \citet{liu2018learning} introduce a virtual root sentence and induce a latent discourse tree via supervised document classification, we generate a root sentence via reconstructing a parent sentence from its children without supervision. 

\section{Experiments}
In this section, we present our experiments for the evalation of the summary generation performance of online reviews. The following section provides the details of the experiments and results. \footnote{The code to reproduce the results is available at: \texttt{https://github.com/misonuma/strsum}}

\subsection{Dataset}
Our experiments use the {\it Amazon product review} dataset \cite{mcauley2015image, he2016ups}, which contains Amazon online reviews and their one-sentence summaries. It includes 142.8 million reviews spanning May 1996 - July 2014. \citet{ma2018hierarchical, wang2018self} used this dataset for the evaluation of their supervised summary generation model. The same domains considered in their previous work are selected for this study; {\it Toys \& Games}, {\it Sports \& Outdoors}, and {\it Movies \& TV}. 

Because our model is trained by identifying and reconstructing a parent sentence from its children, it sometimes fails to construct an appropriate tree for relatively short reviews. It also has a negative influence on summary generation. Therefore, we use reviews with $10$ or more sentences for training, and those with $5$ or more sentences for validation and evaluation. Table \ref{tbl:num} indicates the number of reviews in each domain.

\subsection{Experimental Details}
The source sentences and the summaries share the same vocabularies, which are extracted from the training sources of each domain. We limit a vocabulary to the $50,000$ most frequent words appearing in training sets.

The hyper-parameters are tuned based on the performance using the reference summaries in validation sets. We set $300$-dimensional word embeddings and initialize them with pre-trained FastText vectors \cite{joulin2017bag}. The encoder is a single-layer Bi-GRU with $256$-dimensional hidden states for each direction and the decoder is a uni-directional GRU with $256$-dimensional hidden states.　The damping factor of DiscourseRank is $0.9$. We train the model using Ada-grad with a learning rate of $10^{-1}$, an initial accumulator value of $10^{-1}$, and a batch size of $16$. At the evaluation time, a beam search with a beam size of $10$ is used.

Similar to \cite{see2017get, ma2018hierarchical}, our evaluation metric is the ROUGE-F1 score \cite{lin2004rouge}, computed by the {\it pyrouge} package. We use ROUGE-1, ROUGE-2, and ROUGE-L, which measure the word-overlap, bigram-overlap, and longest common sequence between the reference and generated summaries, respectively.

\begin{table}[t!]
\begin{center}
\scalebox{1.0}{
\begin{tabular}{p{2.9cm}>{\raggedleft}p{1.3cm}>{\raggedleft}p{0.9cm}>{\raggedleft}p{0.9cm}@{}l@{}} \hline
Domains&Train&Valid&Eval&\\ \hline
Toys \& Games&27,037&498&512& \\
Sports \& Outdoors&37,445&511&466& \\
Movies \& TV&408,827&564&512& \\ \hline
\end{tabular}
}
\end{center}
\caption{\label{tbl:num} Number of reviews for training (Train), validation (Valid) and evaluation (Eval).}
\end{table}

\begin{table*}[t!]
\centering
\begin{tabular}{p{4.05cm}>{\raggedleft}p{0.85cm}>{\raggedleft}p{0.85cm}>{\raggedleft}p{0.85cm}|>{\raggedleft}p{0.85cm}>{\raggedleft}p{0.85cm}>{\raggedleft}p{0.85cm}|>{\raggedleft}p{0.85cm}>{\raggedleft}p{0.85cm}>{\raggedleft}p{0.85cm}@{}l@{}} \hline
Domain&\multicolumn{3}{c|}{Toys \& Games}&\multicolumn{3}{c|}{Sports \& Outdoors}&\multicolumn{3}{c}{Movies \& TV}&\\ \hline
Metric&R-1&R-2&R-L&R-1&R-2&R-L&R-1&R-2&R-L&\\ \hline 
\multicolumn{10}{c}{Unuspervised approaches}& \\\hline
TextRank&8.63&1.24&7.26&7.16&0.89&6.39&\bf8.27&\bf1.44&7.35& \\
Opinosis&8.25&1.51&7.52&7.04&1.42&6.45&7.80&1.20&7.11& \\
MeanSum-single&8.12&0.58&7.30&5.42&0.47&4.97&6.96&0.35&6.08& \\
{\em StrSum}&11.61&1.56&11.04&9.15&1.38&8.79&7.38&1.03&6.94& \\
{\em StrSum+DiscourseRank}&\bf11.87&\bf1.63&\bf11.40&\bf9.62&\bf1.58&\bf9.28&8.15&1.33&\bf7.62& \\ \hline
\multicolumn{10}{c}{Supervised baselines}& \\\hline
Seq-Seq&13.50&2.10&13.31&10.69&2.02&10.61&7.71&2.18&7.08& \\
Seq-Seq-att&16.28&3.13&16.13&11.49&2.39&11.47&9.05&2.99&8.46& \\ \hline
\end{tabular}
\caption{\label{tbl:result} ROUGE F1 score of the evaluation set (\%). R-1, R-2 and R-L denote ROUGE-1, ROUGE-2, and ROUGE-L, respectively. The best performing model among unsupervised approaches is shown in \textbf{boldface}. }
\end{table*}

\subsection{Baseline}
For the comparisons, two unsupervised baseline models are employed. A graph-based unsupervised sentence extraction method, {\it TextRank} is employed \cite{mihalcea2004textrank}, where sentence embeddings are used instead of bag-of-words representations, based on \cite{rossiello2017centroid}. As an unsupervised word-level extractive approach, we employ {\it Opinosis} \cite{ganesan2010opinosis}, which detects salient phrases in terms of their redundancy. Because we observe repetitive expressions in the dataset, Opinosis is added as a baseline. Both methods extract or generate a one-sentence summary.

Furthermore, a third, novel unsupervised baseline model {\it MeanSum-single} is introduced, which is an extended version of the unsupervised neural multi-document summarization model \cite{chu2018unsupervised}. While it decodes the mean of multiple document embeddings to generate the summary, MeanSum-single generates a single-document summary by decoding the mean of the sentence embeddings in a document. It learns a language model through reconstruction of each sentence. By comparing with MeanSum-single, we verify that our model focuses on the main review points, and does not simply take the average of the entire document.

As supervised baselines, we employ vanilla neural sequence-to-sequence models for abstractive summarization \cite{hu2015lcsts}, following previous studies \cite{ma2018hierarchical, wang2018self}. We denote the model as {\it Seq-Seq} and that with the attention mechanism as {\it Seq-Seq-att}. The encoder and decoder used are the same as those used in our model.

\begin{figure}[t!]
\begin{center}
\includegraphics[width=7.7cm]{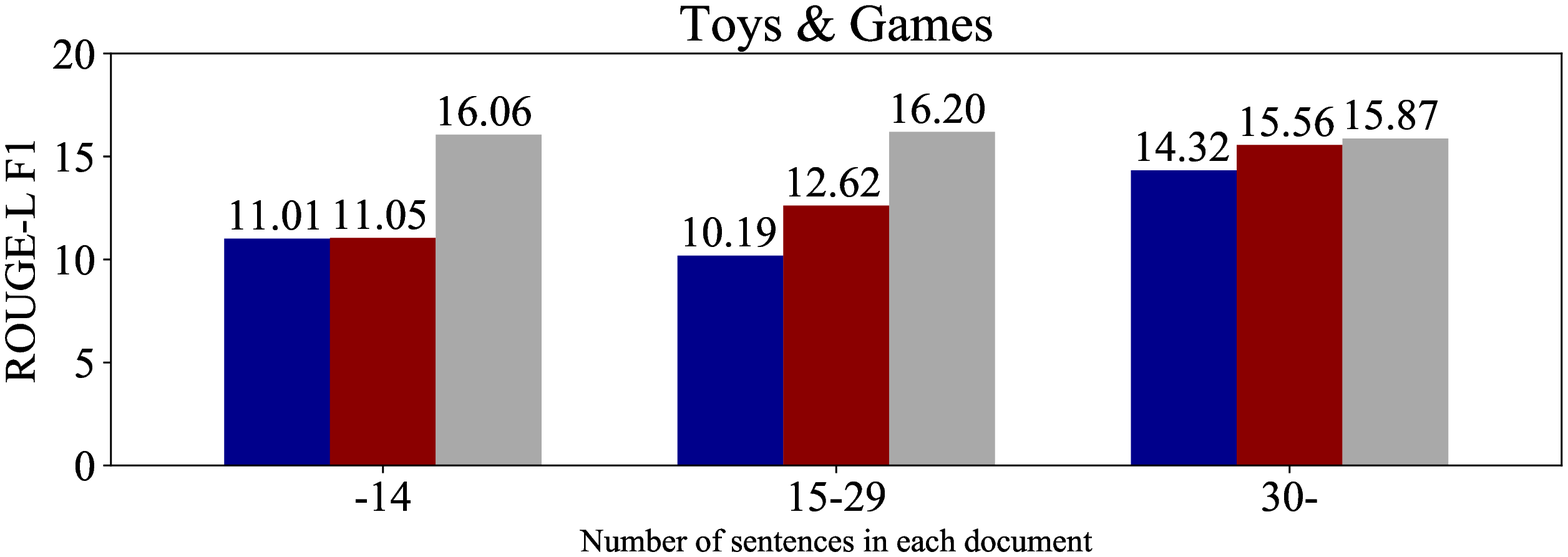}
\end{center}
\begin{center}
\includegraphics[width=7.7cm]{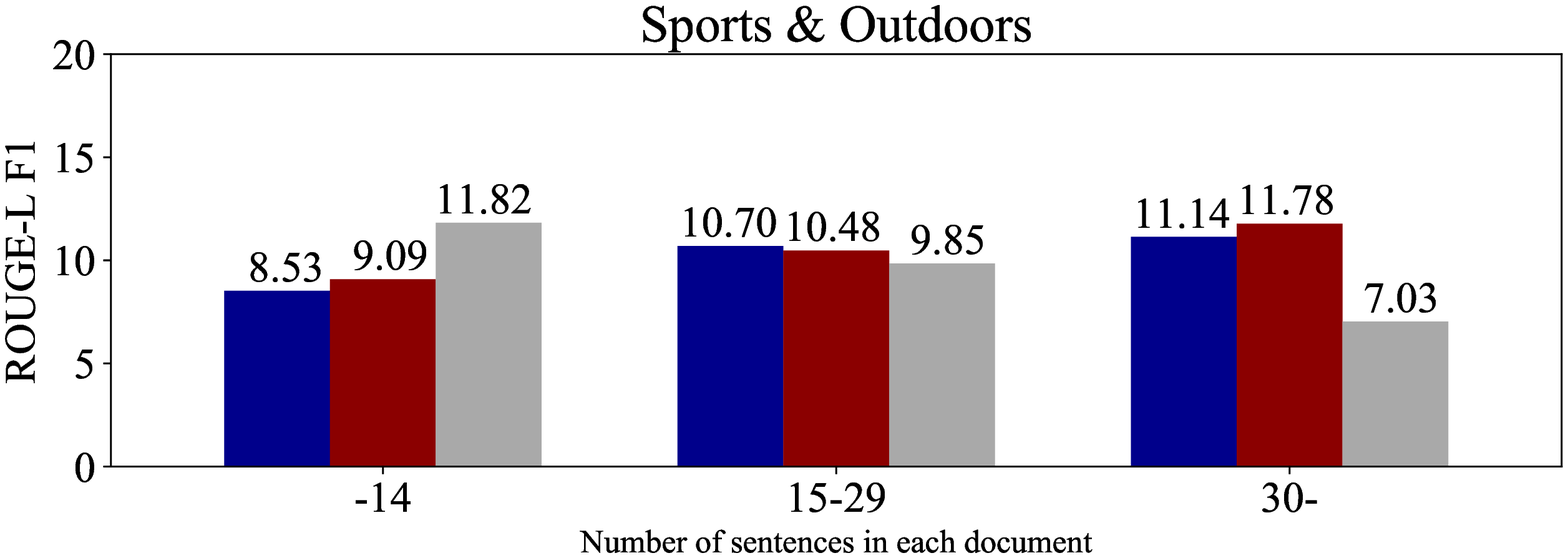}
\end{center}
\begin{center}
\includegraphics[width=7.7cm]{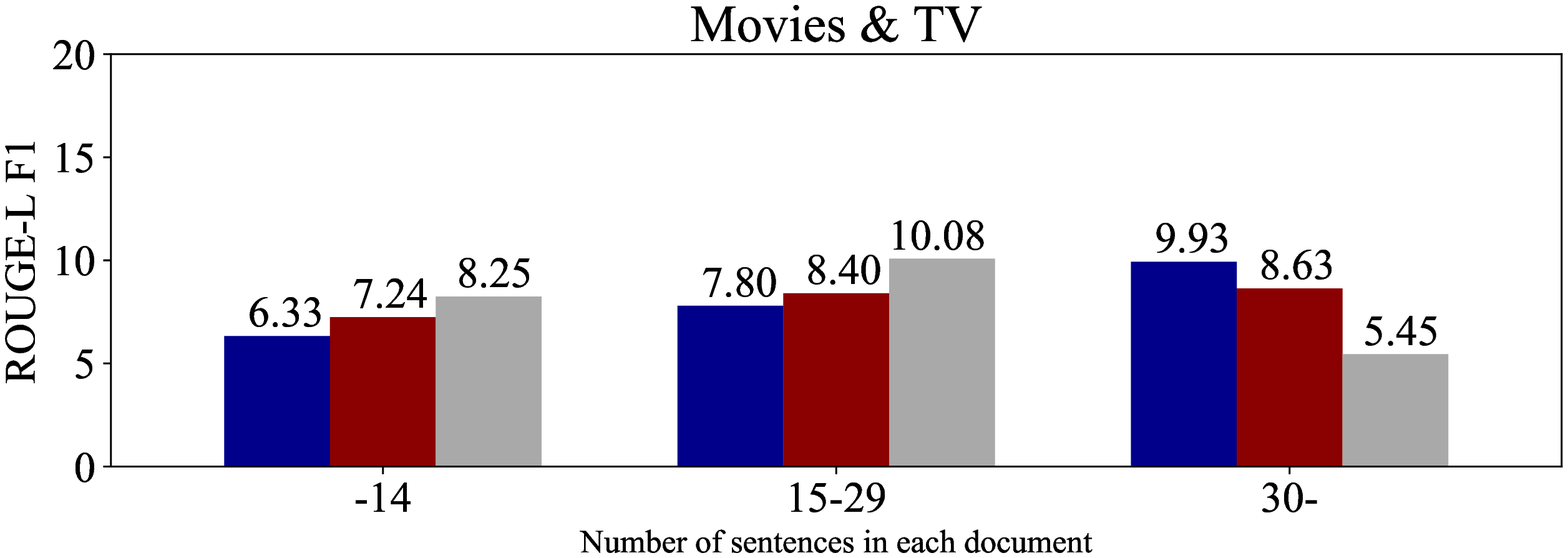}
\end{center}
\begin{center}
\includegraphics[width=7.0cm]{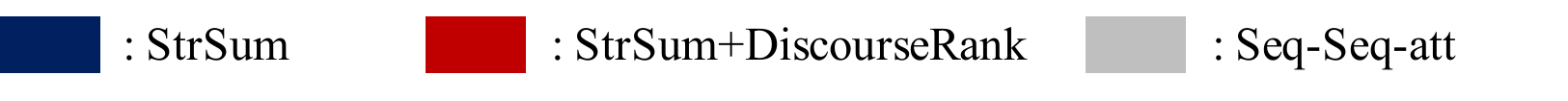}
\end{center}
\caption{ROUGE-L F1 score on evaluation set with various numbers of sentences.}
\label{fig:result_doc_l}
\end{figure}

\begin{figure*}[t!]
\begin{center}
\includegraphics[width=15cm]{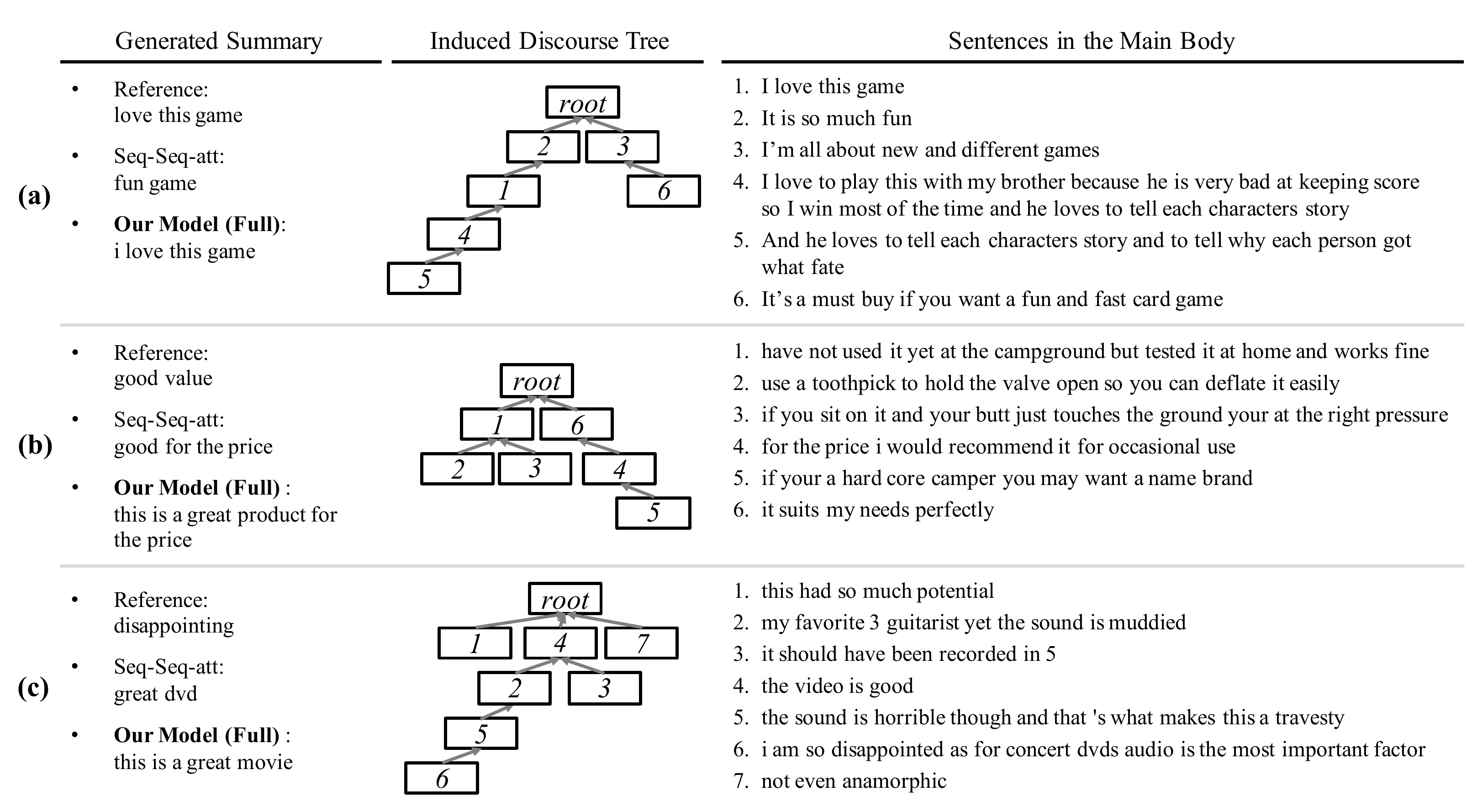}
\end{center}
\caption{Examples of generated summaries and induced latent discourse trees.}
\label{fig:sample}
\end{figure*}


\subsection{Evaluation of Summary Generation}
\label{sec:result}

Table \ref{tbl:result} shows the ROUGE scores of our models and the baselines for the evaluation sets.\footnote{As \citet{yu2016product, ma2018hierarchical} reported, the reviews and their summaries are usually colloquial and contain more noise than news articles. Therefore, the ROUGE scores on the Amazon review dataset are lower than those obtained for other summarization datasets, such as DUC.} With regards to {\em Toys \& Games} and {\em Sports \& Outdoors}, our full model (StrSum + DiscourseRank) achieves the best ROUGE-F1 scores among the unsupervised approaches. As for ROUGE-1 and ROUGE-L, two-tailed t-tests demonstrate that the difference between our models and the others are statistically significant $(p < 0.05)$. Because the abstractive approach generates a concise summary by omitting trivial phrases, it can lead to a better performance than those of the extractive ones. On the other hand, for {\em Movies \& TV}, our model is competitive with other unsupervised extractive approaches; TextRank and Opinosis. One possible explanation is that the summary typically includes named entities, such as the names of characters, actors and directors, which may lead to a better performance of the extractive approaches. For all datasets, our full model outperforms the one using only StrSum. Our models significantly outperform MeanSum-single, indicating that our model focuses on the main review points, and does not simply take the average of the entire document.

Figure \ref{fig:result_doc_l} shows the ROUGE-L F1 scores of our models on the evaluation sets with various numbers of sentences compared to the supervised baseline model (Seq-Seq-att). For the case of a dataset with less than $30$ sentences, the performance of our models is inferior to that of the supervised baseline model. Because our full model generates summaries via learning the latent discourse tree, it sometimes fails to construct a tree, and thus experiences a decline in performance for relatively short reviews. On the other hand, for datasets with the number of sentences exceeding $30$, our model achieves competitive or better performance than the supervised model.

\section{Discussion}  

\subsection{Analysis of the Induced Structure}

Figure \ref{fig:sample} presents the generated summary and the latent discourse tree induced by our full model. We obtained the {\em maximum spanning tree} from the probability distribution of dependency, using Chu--Liu--Edmonds algorithm \cite{chu1965shortest, edmonds1967optimum}. 

Figure \ref{fig:sample}(a) shows the summary and the latent discourse tree for a board game review. Our model generates the summary, {\it "i love this game"}, which is almost identical to the reference. The induced tree shows that the $2$nd sentence {\em elaborates} on the generated summary, while the $3$rd sentence provides its {\em background}. The $4$th and $5$th sentences {\em explain} the $1$st sentence in detail, \emph{i.e.}, describe why the author loves the game.

Figure \ref{fig:sample}(b) shows the summary and latent discourse tree of a camping mattress review. Although there is no word-overlap between the reference and generated summary, our model focuses on the positivity in terms of the {\em price}. On the induced tree, the $1$st to $3$rd sentences provide a {\em background} of the summary and mention the high quality of the product. The $6$th sentence indicates that reviewer is {\em satisfied}, while the $4$th sentence provides its explanation with regards to the {\em price}.

In Figure \ref{fig:sample}(c), we present a failure example of a review of a concert DVD. The reviewer is {\em disappointed} by the poor quality of the {\em sound}; however our model generates a positive summary, {\it "this is a great movie"}. The induced tree shows that the sentences describing the high potential ($1$st), quality of the video ($4$th), and preference to the picture ($7$th), all affect the summary generation. Our model regards the {\em sound} quality as a secondary factor to that of the {\em video}. Therefore, it fails to prioritize the contrasting aspects; the sound and the video, and generates an inappropriate summary. DiscourseRank cannot work well on this example, because the numbers of sentences mentioning each aspect are not significantly different. To solve such a problem, the aspects of each product must be ranked explicitly, such as in \cite{gerani2014abstractive, angelidis2018summarizing}.

\begin{table}[t!]
\begingroup
\begin{center}
\begin{tabular}{p{3.2cm}>{\raggedleft}p{1.5cm}>{\raggedleft}p{1.5cm}@{}l@{}} \hline
Toys \& Games&{\em StrSum}&StrAtt&\\ \hline
Projective&38.58\%&66.07\%& \\
Height&3.06&2.42& \\ \hline
\end{tabular}
\end{center}
\endgroup
\begingroup
\begin{center}
\begin{tabular}{p{3.2cm}>{\raggedleft}p{1.5cm}>{\raggedleft}p{1.5cm}@{}l@{}} \hline
Sports \& Outdoors&{\em StrSum}&StrAtt&\\ \hline
Projective&41.26\%&58.85\%& \\
Height&2.72&2.50& \\ \hline
\end{tabular}
\end{center}
\endgroup
\begingroup
\begin{center}
\begin{tabular}{p{3.2cm}>{\raggedleft}p{1.5cm}>{\raggedleft}p{1.5cm}@{}l@{}} \hline
Movies \& TV&{\em StrSum}&StrAtt&\\ \hline
Projective&36.31\%&61.20\%& \\
Height&3.63&2.37& \\ \hline
\end{tabular}
\end{center}
\endgroup
\caption{\label{tbl:discoursetree} Descriptive statistics for induced latent discourse trees. StrAtt denotes the Structured Attention Model \cite{liu2018learning}.}
\end{table}

Table \ref{tbl:discoursetree} summarizes the characteristics of the induced latent discourse trees. These are compared with those obtained by the Structured Attention model, {\it StrAtt} \cite{liu2018learning}. StrAtt induces single-root trees via the document classification task based on the review ratings. For each domain, our model induces more non-projective trees than StrAtt. Additionally, the height (the average maximum path length from a root to a leaf node) is larger than that of StrAtt. Our model estimates the parent of all the sentences and can induce {\em deeper} trees in which the edges connect trivial sentences. On the other hand, StrAtt identifies salient sentences required for the document classification, and thus induces {\em shallow} trees that connect the salient sentences and others. As our model prevents the summary from focusing on trivial or redundant sentences by inducing deep and complex trees, it specifically achieves higher performance when considering relatively long reviews.

\begin{figure}[t!]
\setlength\textfloatsep{0pt}
\begin{flushleft}
\includegraphics[width=7.2cm]{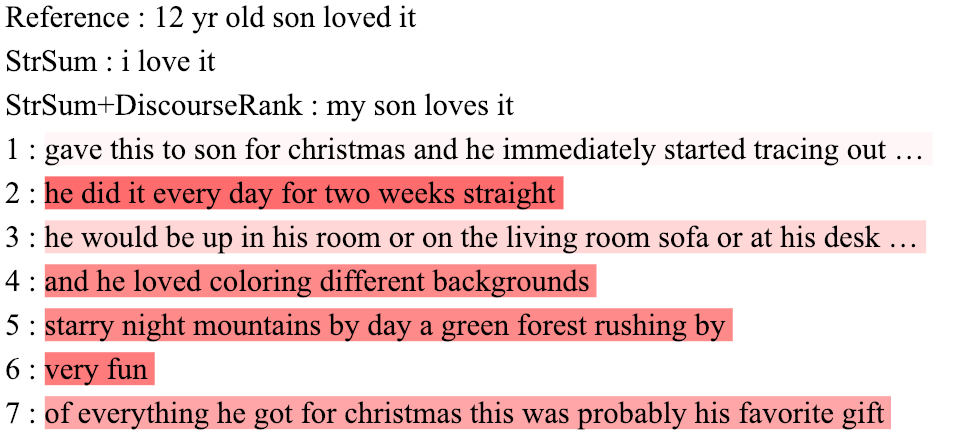}
\end{flushleft}
\begin{center}
\vspace{-5pt}
\includegraphics[width=0.7cm]{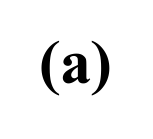}
\end{center}
\begin{flushleft}
\vspace{-17pt}
\includegraphics[width=7.7cm]{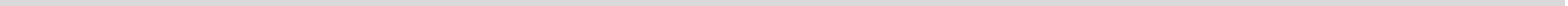}
\end{flushleft}
\begin{flushleft}
\includegraphics[width=7.7cm]{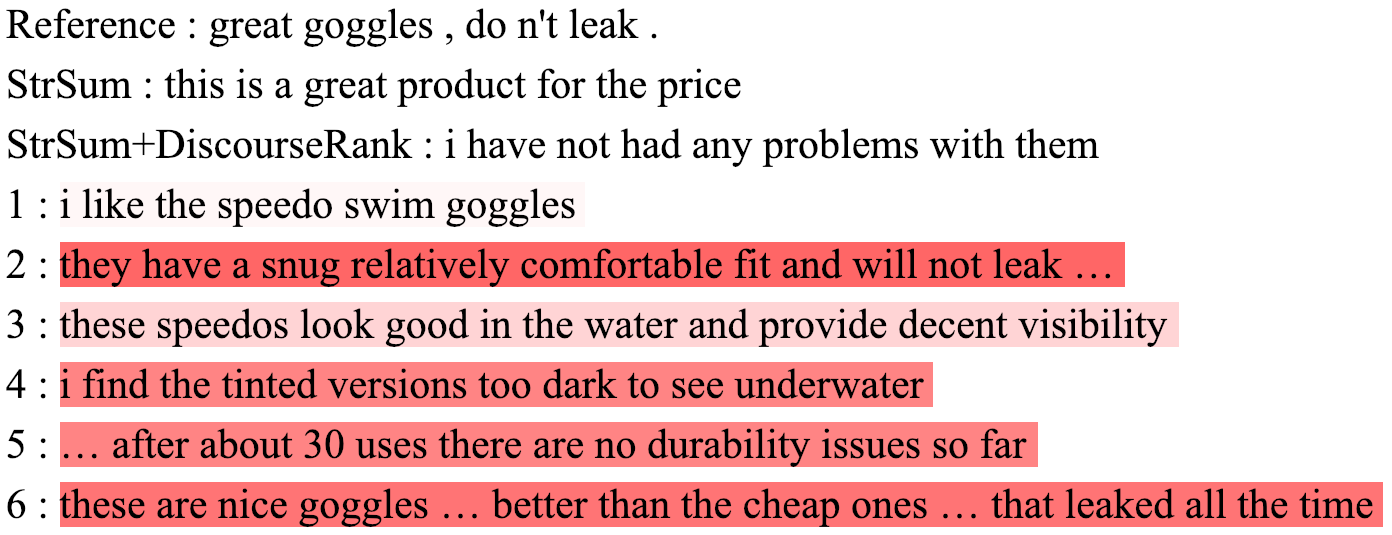}
\end{flushleft}
\begin{center}
\vspace{-5pt}
\includegraphics[width=0.7cm]{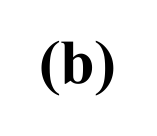}
\end{center}
\caption{Visualization of DiscourseRank. The darker the highlightning, the higher the rank score. The references and generated summaries are also shown.}
\label{fig:eigen}
\end{figure}

\subsection{DiscourseRank Analysis}

In this section, we demonstrate how DiscourseRank affects the summary generation. Figure \ref{fig:eigen} visualizes the sentences in the main body and their DiscourseRank scores. We highlight the sentences that achieve a high DiscourseRank score with a darker color. 

A review of a car coloring book is presented in Figure \ref{fig:eigen}(a). As expected, the score of the $1$st sentence is low, which is not related to the review evaluations, that is, DiscourseRank emphasizes the {\em evaluative} sentences, such as the $2$nd and $6$th sentences.

A review of swimming goggles is presented in Figure \ref{fig:eigen}(b). The reviewer is satisfied with the quality of the product. The highlighting shows that DiscourseRank focuses on the sentences that mention leaking (\emph{e.g.}, the $2$nd and $5$th). While our model (with only StrSum) emphasizes the {\em price} sufficiency, DiscourseRank generates a summary describing that there is no issue with the {\em quality}.

\section{Conclusion}
  
In this work, we proposed a novel unsupervised end-to-end model to generate an abstractive summary of a single product review while inducing a latent discourse tree. The experimental results demonstrated that our model is competitive with or outperforms other unsupervised approaches. In particular, for relatively long reviews, our model achieved competitive or better performance compared to supervised models. The induced tree shows that the child sentences present additional information about their parent, and the generated summary abstracts the entire review.

Our model can also be applied to other applications, such as argument mining, because arguments typically have the same discourse structure as reviews. Our model can not only generates the summary but also identifies the argumentative structures. Unfortunately, we cannot directly compare our induced trees with the output of a discourse parser, which typically splits sentences into elementary discourse units. In future work, we will make comparisons with those of a human-annotated dataset.

\section*{Acknowledgments}

We would like to thank anonymous reviewers and members of the Sakata\&Mori Laboratory at the Graduate School of Engineering for their valuable feedback. This work was supported by CREST, JST, the New Energy and Industrial Technology Development Organization (NEDO) and Deloitte Tohmatsu Financial Advisory LLC.





\onecolumn

\section*{Supplemental Material}
\label{sec:supplemental}

\begin{figure*}[t!]
\begin{center}
\includegraphics[width=16cm]{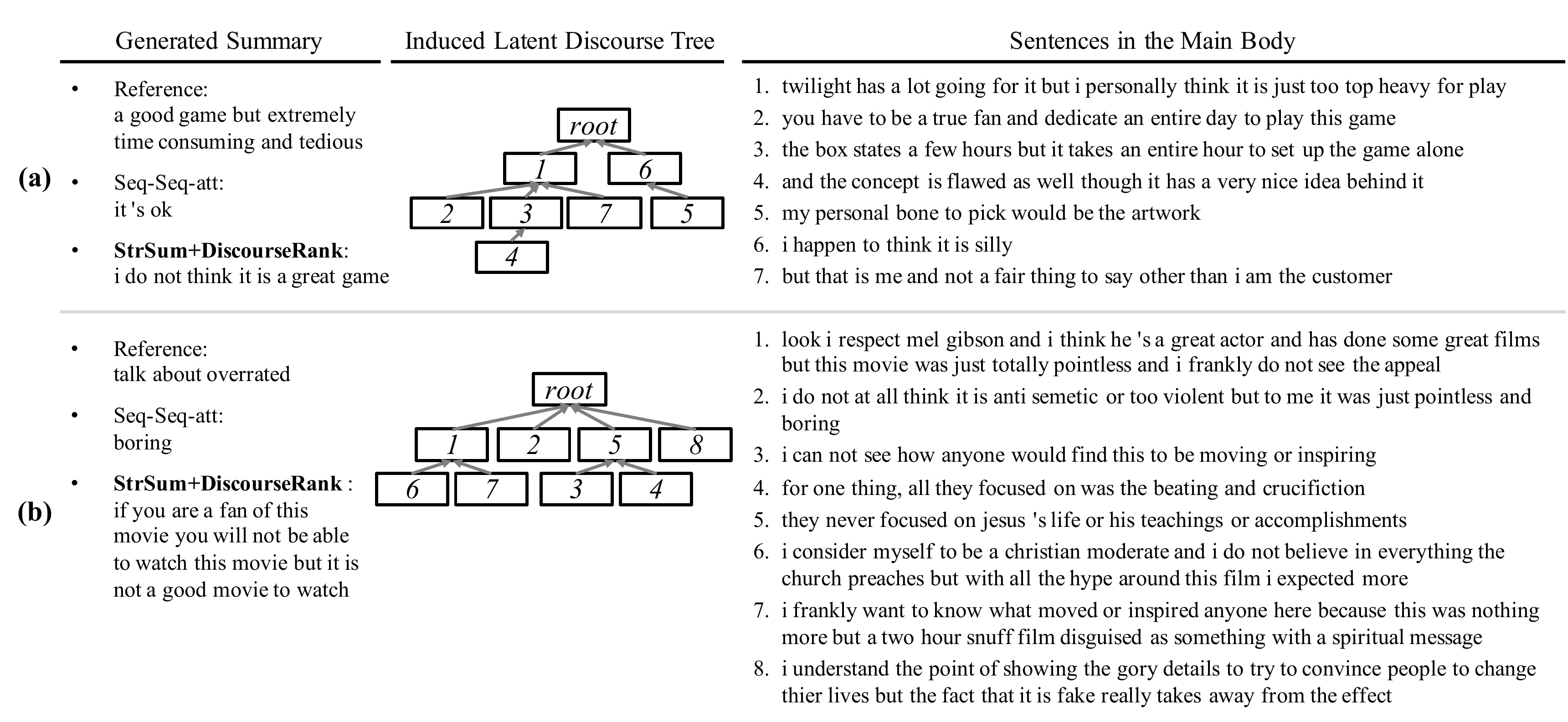}
\end{center}
\caption{Examples of generated summaries and induced latent discourse trees for {\em negative} reviews. (a) shows a board game review. The induced tree shows that the $1$st and $6$th sentences present additional information about the generated summary. While the $1$st to $4$th indicate the {\em heaviness} of the game, the $5$th and $6$th criticize the {\em artwork}. The $2$nd, $3$rd, and $4$th present the additional information about the parent. (b) presents a movie review. The $1$st and $2$nd sentences describe the whole evaluation, while $6$th and $7$th {\em strengthen} the opinion. The $3$rd to $5$th mention the boring points in {\em detail}. Although our model catches the negativeness, the summary is redundant probably because each sentence in the body is relatively long.
}
\label{fig:negativesamples}
\end{figure*}

\begin{figure*}[t!]
\begin{center}
\includegraphics[width=16cm]{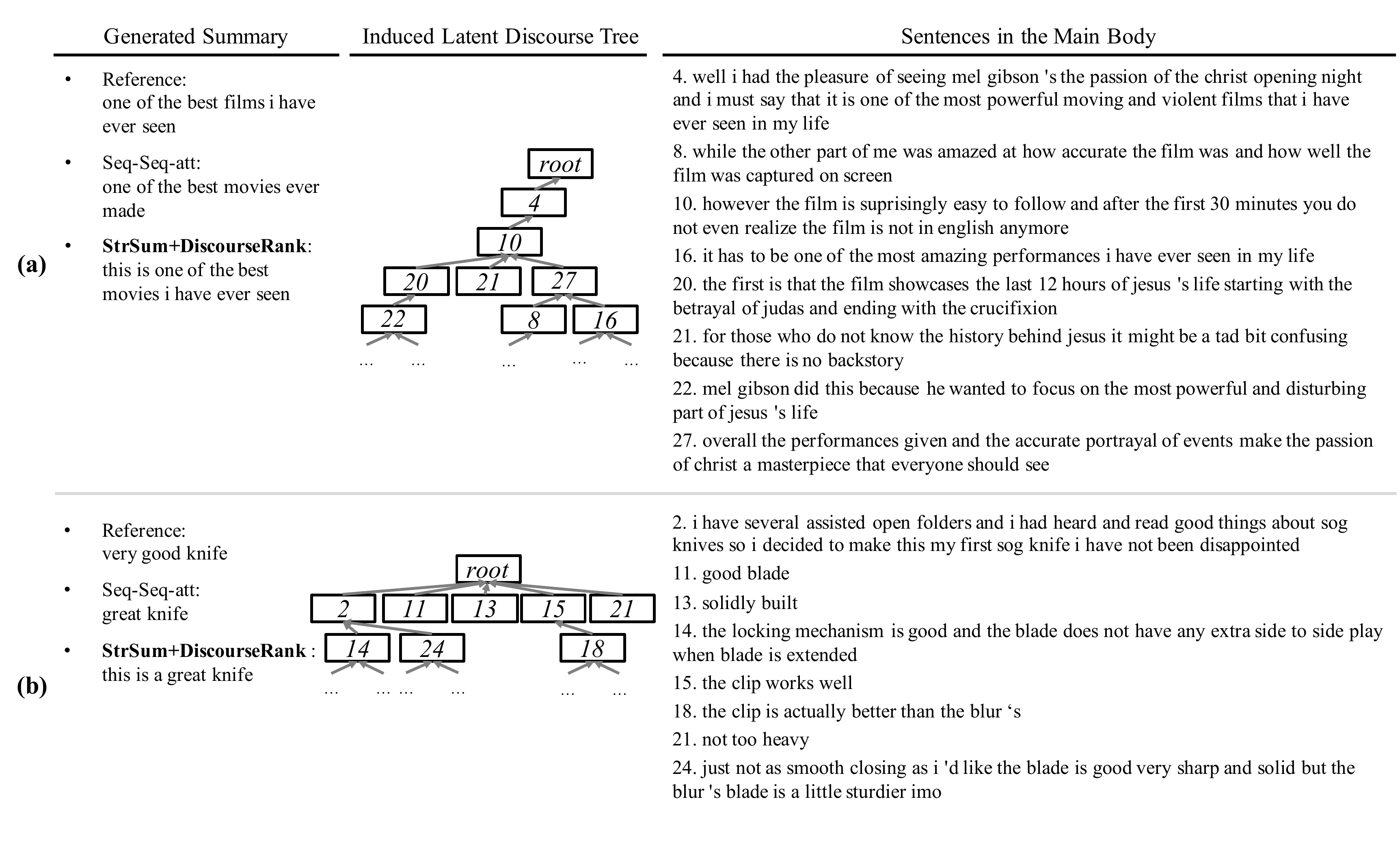}
\end{center}
\caption{Examples of generated summaries and induced latent discourse trees for {\em long} reviews. (a) shows a movie review. The $4$th sentence mentions the whole positiveness. The $10$th describes that the contents are {\em easy to follow}, while the $20$th to $22$nd show the {\em detail} of the contents. The $27$th mentions the {\em performance} and {\em accurate portrayal}, and the $8$th and $16$th {\em elaborate} on the latter and the former, respectively. (b) presents a pocket knife review. The $11$th, $13$th, $15$th, and $21$st sentences concisely describe the goodness in each {\em aspect}. The $14$th, $24$th, and $28$th {\em elaborate} on the parents.}
\label{fig:longsamples}
\end{figure*}

This supplemental material provides examples of the generated summaries and the latent discourse trees induced by our model. Figure \ref{fig:negativesamples} and Figure \ref{fig:longsamples} show samples of {\em negative} reviews and relatively {\em long} reviews, respectively. We present them with comparisons of the reference and generated summaries by the supervised neural sequence-to-sequence model with attention mechanism (Seq-Seq-att). 

\newpage

\end{document}